\pdfoutput=1
\documentclass{article}

\usepackage[dblblindworkshop,final]{neurips_2026}
\workshoptitle{Foundation Models for Temporal Systems: From Forecasting to World Modeling}

\usepackage[utf8]{inputenc}
\usepackage[T1]{fontenc}
\usepackage{hyperref}
\usepackage{url}
\usepackage{booktabs}
\usepackage{amsfonts}
\usepackage{nicefrac}
\usepackage{microtype}
\usepackage{xcolor}
\usepackage{graphicx}
\hypersetup{pdftitle={Do Your Own Research: Learning to Forecast by Learning to Search},
  pdfauthor={Yusuf Afifi, Artur Kiulian, Anton Polishko, Mykola Khandoga, Hamudi Naanaa, Alina Krasnobrizha}}

\usepackage{enumitem}
\title{Do Your Own Research:\\Learning to Forecast by Learning to Search}

\author{%
  Yusuf Afifi\thanks{Corresponding author.} \quad Artur Kiulian \quad Anton Polishko \\
  \textbf{Mykola Khandoga} \quad \textbf{Hamudi Naanaa} \quad \textbf{Alina Krasnobrizha} \\
  Future Principle \\
  \texttt{yusuf@futureprinciple.com}
}

\begin{document}

\maketitle
\setcounter{footnote}{0}

\begin{abstract}
Outcome-based reinforcement learning can train language models to
forecast real-world events, but prior forecasting work either freezes research
context before training or deploys agentic research only at test time,
so the skill of \emph{gathering} evidence is never shaped by the
reward. We introduce an agentic forecasting environment, dataset, and harness
built from 2{,}100+ resolved Polymarket questions; the agent
acquires its own context at rollout time (web search, page reading, and
financial time series, all restricted by layered leak filtering to
information published before each question's cutoff), and we train
Qwen3.5-35B-A3B (3B active parameters) on it with single-epoch GRPO
under a Brier-score reward. Training changes how the agent interacts with information:
calibration improves 30--40\%, and search attempts fall from 3.8 to
2.25 per rollout as evidence discipline is learned. Evaluated in an
identical harness against four frontier models, the trained policy
also finishes ahead of every frontier model tested at evidence-based
forecasting, including Claude Opus~4.5 (soft-Brier 0.254 vs.\ 0.256,
$n{=}265$), at about 5\% of the inference cost, and its margin is
widest on the hardest questions, the ones the crowd itself had not
decided.
We release the environment, dataset, and per-rollout records as a
reusable harness for temporal forecasting agents.\footnote{\url{https://github.com/afifi-yusuf/prime-forecast}}
\end{abstract}

\section{Introduction}

Prediction markets such as Polymarket let traders buy and sell
contracts on real-world events; every question therefore carries a
continuously updated crowd probability (the market price) and, once
the event resolves, a verified outcome; few other domains give agentic
RL verifiable, dated answers at this scale. Two recent lines of work showed that outcome-based RL on these questions produces
strong forecasters. \citet{turtel2025} train a 14B model with
outcome-only RL on 10k resolved questions plus 100k synthetic ones,
reaching statistical parity with o1 on accuracy while halving its
calibration error;
\citet{mantic2025} fine-tune gpt-oss-120b on ${\sim}10$k questions to a
Metaculus baseline score marginally above Gemini~3~Pro, and find that research
context drives most of the training gain (+7 points with it, +3
without). Both freeze the context before training, pasted headlines in the
former, a pre-generated ``research phase'' in the latter, leaving results
bounded by the quality of that one-time collection. \citet{turtel2025} attribute part
of their remaining gap to the information lag of frozen headlines, and
\citet{mantic2025} acknowledge that frozen context caps performance, since no policy
improvement can recover information the prompt lacks, naming the
follow-up outright: ``Folding
retrieval into the training loop is the natural way to test this.'' On the
other hand, \citet{murphy2026} makes forecasting agentic at \emph{test
time} by introducing a multi-turn harness in which a frozen model
sequentially updates an explicit belief state through iterative tool
use. Even optimized retrieval scaffolds, however, only approach crowd
accuracy~\citep{halawi2024}; without non-public information no scaffold
reliably beats the market, hence we focus on
\emph{evidence-based forecasting}, where the market price is not given
to the models. 

We close the frozen-context gap by folding a multi-turn
search-and-forecast scaffold directly into the RL training loop; RL with verifiable rewards has already
incentivized reasoning, most prominently in DeepSeek-R1~\citep{guo2025},
and search, as in Search-R1~\citep{jin2025}; the enabling step for
forecasting is making it temporally sound on resolved historical
questions. At rollout time, the agent decides, over multiple turns,
what to search, which pages to read, and which time series to pull,
with layered leak filtering restricting every channel to information
published before the question's cutoff. Bringing context acquisition into the RL loop under the same verifiable reward moves the field from optimizing a single-turn output to training a
multi-turn reasoning forecaster, allowing the agent to improve its
search queries, weigh how far to trust each result, and learn which
evidence bears on the outcome. In contrast to time-series foundation models~\citep{ansari2024,das2024}, which forecast from a fixed numeric context window, our agent chooses its own context at rollout time, numeric series included, and context acquisition itself becomes the learned temporal skill.
Our contributions:
\begin{enumerate}[leftmargin=*,itemsep=1pt,topsep=1pt]
  \item \textbf{An agentic forecasting dataset, environment, and
    harness} (open-sourced on the Prime Intellect Environments Hub): 2{,}113 training
    and 265 held-out resolved Polymarket questions with strict temporal
    eligibility, as well as the harness configuration and code to allow multi-turn agentic forecasting in both training and testing.
  \item \textbf{An identical-harness evaluation of interactive
    forecasting agents}: trained, untrained, and four frontier models
    run with the same tools, filters, and turn budget, with
    platform-served policies, per-rollout capture, and trace-level
    audits as part of the method.
  \item \textbf{Results and benchmarks}: single-epoch GRPO takes an open 35B-A3B
    model ahead of every frontier model tested at evidence-based
    forecasting, beating Claude Opus~4.5, Sonnet, Gemini~Pro, and
    Flash at about 5\% of the
    inference cost, with the widest margins on
    uncertain questions.
\end{enumerate}

\section{The environment and dataset}
\label{sec:env}

\paragraph{Task and harness.} Each episode presents one resolved binary
Polymarket question with its resolution criteria and a forecast cutoff
date; the agent must produce a probability that the event occurs, using
only information available before the cutoff. It runs a multi-turn tool
loop (up to 10 turns): web search (${\le}3$ calls, served through an
Amazon Bedrock AgentCore gateway), URL reading with LLM
summarization, financial and economic time
series truncated at the cutoff, revision-dated Wikipedia access, a trend-extrapolation tool, and a
terminal \texttt{submit} action returning a probability and rationale.
The system prompt requests an explicit belief state~\citep{murphy2026}: every tool call
is meant to carry an updated probability and evidence lists, and each prediction
has a human-readable reasoning trace, an auditable update trajectory
from prior to final probability, and an inspectable world-state model.

\paragraph{Leak filtering.} Because questions are historical, the
central validity threat is information from after the cutoff leaking
into the rollout. Retrieved content therefore passes three layers
before the model sees it: a domain blocklist (prediction-market sites
and known mirrors are never fetched); a heuristic date filter that
parses publication dates, drops post-cutoff results, and scrubs lines
containing post-cutoff facts; and a per-result keep/drop judgment by Claude Haiku (via
Amazon Bedrock) against the question and cutoff, which catches undated
pages and indirect outcome reveals. Haiku alone scored perfectly among
five judges benchmarked on labeled keep/drop cases (the others:
Gemma-3-12B, gpt-oss-20b, gpt-oss-safeguard-20b, and Nova~2~Lite); every alternative erred toward over-dropping valid
evidence rather than leaking outcomes (per-model results in the
repository).

\paragraph{Dataset.} Questions are drawn from the public Polymarket
APIs and filtered for liquidity (volume ${\ge}\$5{,}000$), genuine
undecidedness (market price at the cutoff in $[0.10, 0.90]$: the crowd
had not effectively settled the question), and forecast horizon
(cutoffs drawn hash-stably 7--90 days before
resolution). Every cutoff and resolution postdates the base model's release window, precluding any outcome from entering its parametric knowledge; test questions resolve strictly after all training questions for backtesting compatibility. The training split has
2{,}113 questions (politics 40\%, crypto/finance 34\%, the rest
weather, AI/tech, and other); the held-out test split has 265.

\paragraph{Reward and objective.} A submitted probability $p$ on outcome
$y \in \{0,1\}$ earns $r = 1 - (p-y)^2$, one minus the Brier
score~\citep{brier1950}. Prior work optimizes this reward against pre-collected context $c$,
maximizing
$J(\theta) = \mathbb{E}_{(q,c,y)\sim\mathcal{D}}\,
\mathbb{E}_{\hat{p}\sim\pi_\theta(\cdot\mid q,c)}
[1-(\hat{p}-y)^2]$~\citep{mantic2025}; here context is produced inside
the rollout: the policy generates a multi-turn trajectory $\tau$ of
tool calls in the leak-filtered pre-cutoff environment
$\mathcal{E}_{<t(q)}$ and is scored on the probability
$\hat{p}_\tau$ it submits,
\begin{equation}
J(\theta) = \mathbb{E}_{(q,\,y)\sim\mathcal{D}}\;
\mathbb{E}_{\tau\sim\pi_\theta \otimes\, \mathcal{E}_{<t(q)}}
\left[\,1-(\hat{p}_\tau-y)^2\,\right].
\end{equation}
Non-submission earns $0.55$, low enough that finishing dominates
stalling. The Brier score is strictly
proper~\citep{savage1971,gneiting2007}: truthful probabilities are
uniquely optimal; it is bounded, giving low-variance policy
gradients~\citep{mantic2025}; and it matches prior
conventions~\citep{turtel2025}. The policy is trained with the
market-price tools hidden; its only paths to reward are research and
priors. The environment is built on
\texttt{verifiers}; training is
single-epoch Group Relative Policy Optimization (GRPO)~\citep{shao2024} with LoRA
(33 steps $\times$ 64 questions $\times$ 8 rollouts) on
Qwen3.5-35B-A3B, a sparse mixture-of-experts model with 3B active
parameters, via \texttt{prime-rl} on Prime Intellect's stack.

\section{Evaluation setup}
\label{sec:eval}

All policies (trained, untrained base, Claude Opus~4.5, Claude
Sonnet~4.5, Gemini~3.1~Pro, and Gemini~3.6~Flash) run the
\emph{identical} environment: same tools, same leak filters, same turn
and search budgets. The Qwen policies are served by the training
platform's own inference stack and captured per-rollout via a results
webhook; frontier models run the same environment
through an API proxy.

We report soft-Brier, the mean Brier score with $0.25$ (the score of an
uninformative $0.5$ forecast) imputed whenever a policy declines to
answer, and expected calibration error (ECE) over submitted forecasts with
ten equal-mass bins (the gap between stated confidence and empirical
frequency), following \citet{turtel2025}; both are lower-is-better. All
evaluations use live retrieval with no access to the market's own
price; we call this \emph{evidence-based forecasting}. Price access makes every model, from the untrained 35B to Opus~4.5, converge on the crowd without reliably beating it, so a market-visible comparison measures price-copying, not forecasting (\citealp{halawi2024}; Appendix~\ref{app:market}).

\section{Results}
\label{sec:results}

\begin{figure}[t]
  \centering
  \includegraphics[width=0.8\linewidth]{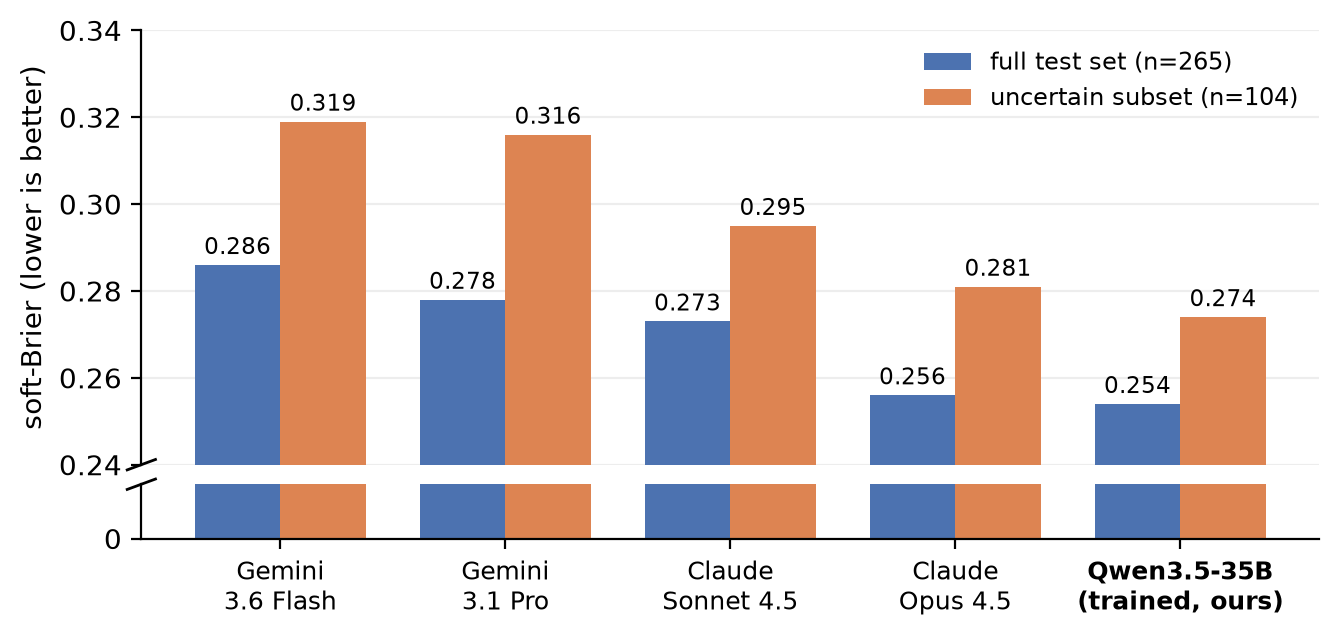}
  \caption{Soft-Brier on the full held-out test split ($n{=}265$) and
  on the pre-declared uncertain subset (cutoff price in $[0.30,0.70]$,
  $n{=}104$). The range $[0,0.24]$ is compressed for readability.}
  \label{fig:main}
\end{figure}

\begin{figure}[t]
  \centering
  \includegraphics[width=1.0\linewidth]{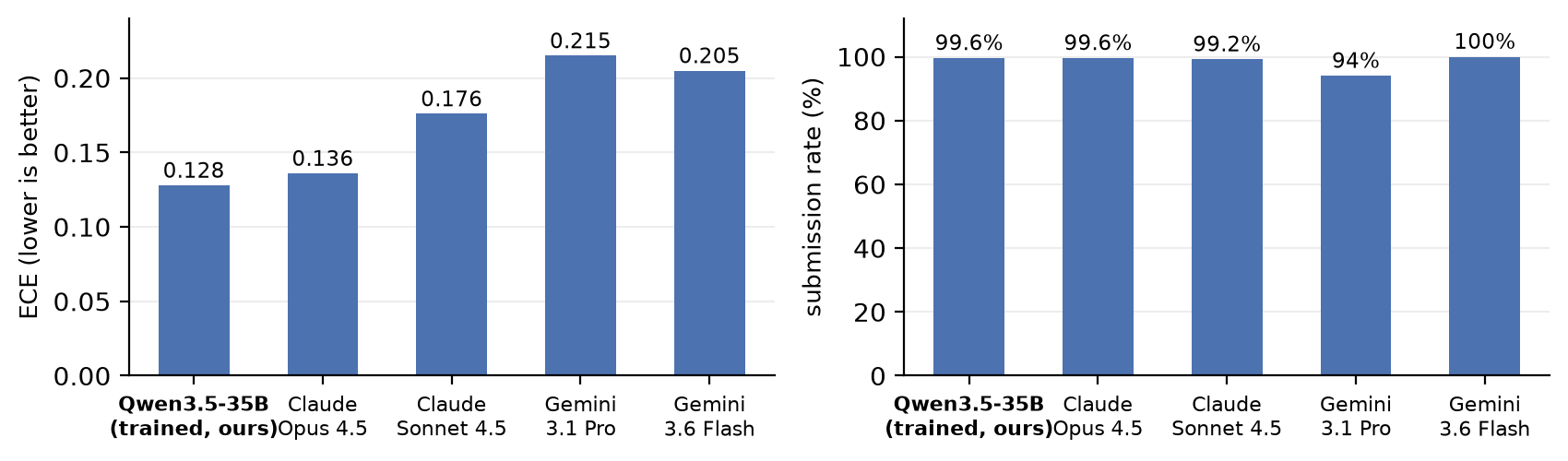}
  \caption{ECE over submitted forecasts and submission rate on the
  held-out test split ($n{=}265$).}
  \label{fig:quality}
\end{figure}

\paragraph{Beating the frontier.}
Figure~\ref{fig:main} gives the headline
result: the RL-trained policy finishes ahead of every frontier model
tested, edging Claude Opus~4.5 (0.254 vs.\ 0.256) and leading Sonnet,
Gemini~Pro, and Flash by 0.019--0.032. A 3B-active open
model, trained for one epoch on ${\sim}2{,}100$ questions, tops the
frontier at about 5\% of the inference
cost.\footnote{Rollouts average ${\sim}36$k input / ${\sim}2.2$k output
tokens per question regardless of policy; at August 2026 list prices
(Opus~4.5 \$5/\$25 per million tokens, Qwen3.5-35B-A3B \$0.25/\$0.75
on our stack), ${\sim}\$0.24$ vs.\ ${\sim}\$0.011$ per question.} The
trained model also posts the lowest calibration error of the panel
(Figure~\ref{fig:quality}); submission rates are 99--100\% throughout,
except Gemini~Pro at 94\%. 

\paragraph{Margins widen in low-confidence markets.}
We pre-declared a subset of $n{=}104$ questions with a cutoff price in $[0.30, 0.70]$, those the crowd itself had not yet decided, where the trained advantage concentrates (Figure~\ref{fig:main}). While every score is higher on this harder subset (the crowd itself scores 0.232), the margins widen. The trained policy (0.274) leads Opus by 0.007 (0.281), Sonnet by 0.021 (0.295), and Pro and Flash by 0.042--0.045 (0.316, 0.319); every paired per-question gap is larger than on the full set, and the Opus gap grows from a near-tie to 0.007. When the outcome is obvious, every competent
researcher finds it and the forecasters converge; when the market is
uncertain, forecasts rest on each agent's own research and
intuition, and the benefit of training is reaped.

\begin{figure}[t!]
  \centering
  \includegraphics[width=1.0\linewidth]{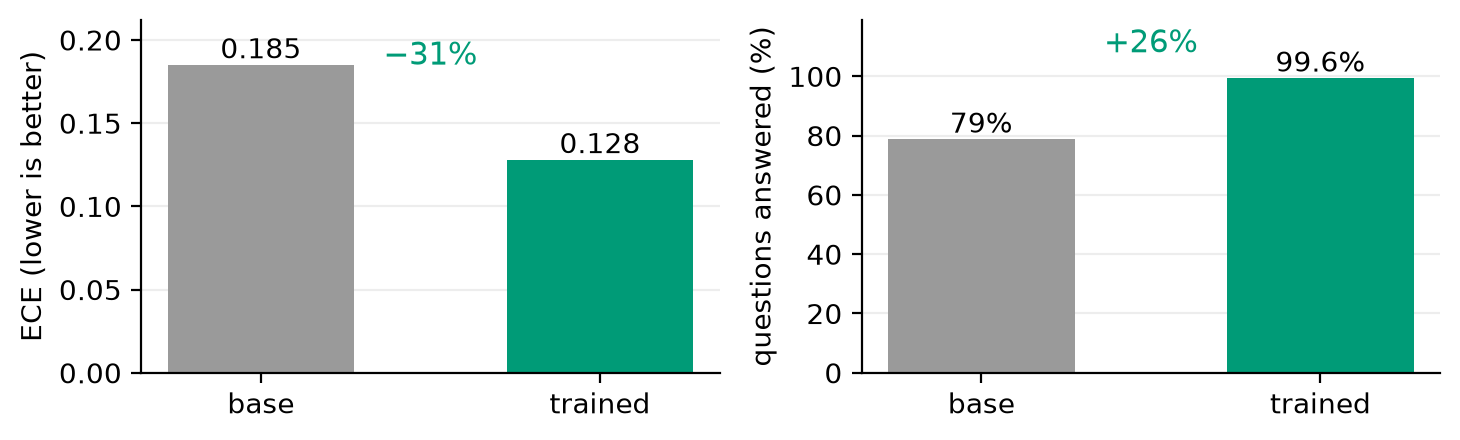}
  \caption{Comparing calibration and coverage of base and trained
  Qwen3.5-35B-A3B.}
  \label{fig:behavior}
\end{figure}

\paragraph{Calibration and coverage improve.}
\label{sec:behavior}
Training improves soft-Brier over the untrained base in every setting we measured, though its largest and most reliable gains are in calibration and coverage (submission rate; Figure~\ref{fig:behavior}). ECE falls 0.185${\to}$0.128 ($-31\%$) in the evidence-based setting, placing the trained model ahead of the frontier panel (Figure~\ref{fig:quality}), while coverage rises from 79\% to ${\sim}100\%$ ($+26\%$). Search attempts also fall from 3.8 to 2.25 per rollout: the trained policy re-queries less after empty or filtered results and reasons more from the surviving evidence. This 30--40\% calibration improvement holds in every train/eval pair of the repository's variant runs (0.099${\to}$0.065 with the market price visible; 0.170${\to}$0.103 in a search-off, no-market run). Rollout traces show this discipline directly (examples in Appendix~\ref{app:traces}).

\section{Conclusion}

While outcome-based RL in prediction-market environments has evidently improved the forecasting priors of models, previous approaches are scaffold-dependent and context-sensitive. Internalizing context acquisition, training the act of research itself, removes this extrinsic sensitivity: agents become intrinsically better forecasters. By teaching a 35B model how to conduct its own research, we elevate both its parametric accuracy and behavioral discipline to frontier levels, establishing a more representative and fair way to train and evaluate forecasting agents. In short, the model learns to forecast by learning to search, and, just as importantly, when not to: search attempts fall from 3.8 to 2.25 per rollout even as accuracy rises and calibration error falls 31\%. Future work should explore scaling this environment beyond binary prediction markets to categorical outcomes, as well as expanding the action space to allow agents to run their own sandboxed quantitative models.

\begin{ack}
We thank Prime Intellect for credits on its hosted training service, for hosting our environment on its Environments Hub, and for the open-source \texttt{verifiers} and \texttt{prime-rl} libraries our training stack builds on. We also thank Amazon Web Services and Google Cloud for the credits that funded the frontier-model evaluations, and Sergey Knysh for guidance and mentorship throughout the project.
\end{ack}

\appendix

\section{Why the market price is withheld}
\label{app:market}
\begin{figure}[h]
  \centering
  \includegraphics[width=0.8\linewidth]{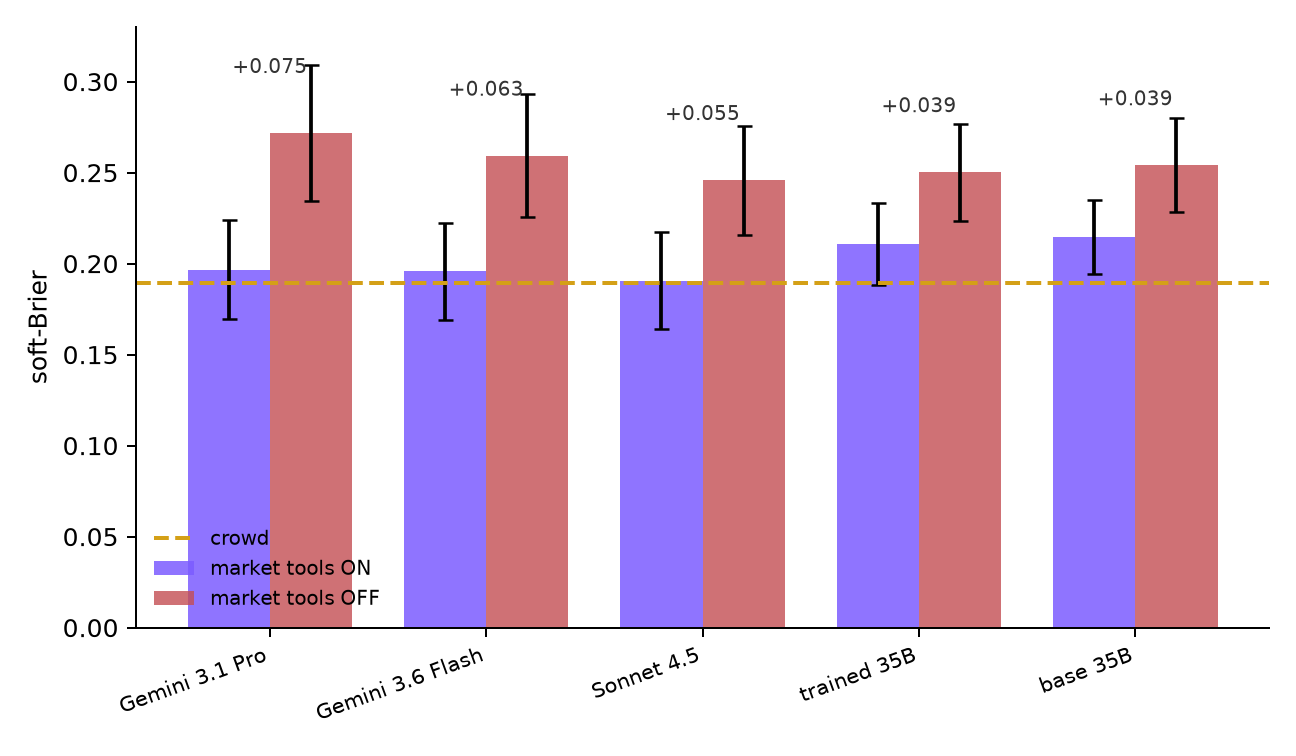}
  \caption{Anchor-worth: the paired Brier cost of withholding the
  market price, per policy. Frontier models gain most from the crowd
  anchor.}
  \label{fig:anchor}
\end{figure}

With market-price tools available, every policy converges toward the
crowd (soft-Brier band 0.186--0.215 vs.\ crowd 0.189) and none
reliably beats it. Withholding the price costs each policy its \emph{anchor-worth},
the paired Brier cost of losing the crowd's probability, measured on
the same questions: ${+}0.055$ to ${+}0.075$ for the frontier models and
${+}0.04$ for the 35B policies (trained and base), several times any
training effect we measured.
Comparisons between agents can therefore be reversed by choosing who
sees the price (Figure~\ref{fig:anchor}).

\section{Example rollouts}
\label{app:traces}
Two complete rollouts on the same question, verbatim from the archived records; long tool
outputs are shown as their leading results with elisions marked
[\dots]. $\triangleright$ marks a tool call (with the belief state the agent
attached to it), $\triangleleft$ the harness response.
{\small

\subsection*{Trained policy: ``Will Donald Trump have a diplomatic meeting with Iran by June 30?''}
Cutoff 2026-06-11; resolved NO. Submitted $0.20$; Brier $0.04$.
Captured from the market-visible variant evaluation; the market tools
were available and never called.

\paragraph{Turn 1.}
\emph{I need to research whether Donald Trump will have a diplomatic meeting with Iran by June 30, 2026. This is a political/foreign policy question about potential diplomatic relations between the US and Iran. First, I should search for information about Trump's foreign policy stance on Iran, current US-Iran relations, and any diplomatic initiatives or negotiations that might be happening before my cutoff date. I need to be careful not to use outcome words in my search to avoid post-cutoff information. Let me start with a web search to understand the current state of US-Iran relations and Trump's potential foreign policy approach.}

\noindent$\triangleright$ \texttt{web\_search}(``Donald Trump Iran diplomatic policy nuclear deal''); belief $p{=}0.15$ (low confidence); ``Starting with historical base rate - Trump's prior term ended US relations with Iran over nuclear deal, making diplomatic meetings unlikely absent major political changes. No evidence yet of current initiatives.''; for: Trump previously withdrew from JCPOA in 2018; US-Iran relations remain severely strained; against: None identified yet; uncertain: Future political developments; Any shift in Trump's foreign policy approach

\noindent$\triangleleft$ 2 result(s): \emph{United States withdrawal from the Joint Comprehensive Plan of Action} (en.wikipedia.org): ``The Trump administration issued certifications in April and July 2017 that Iran was complying with the deal. [ 23 ] [ 24 ] But in October 2017, Trump announced that the US would not make the certification provided for un [\dots]'' \quad \emph{What Is the Iran Nuclear Deal?} (cfr.org): ``Diplomacy to revive this arms control agreement has faced multiple stumbling blocks, including Iran's nuclear advances and its links to conflicts in the Gaza Strip and Ukraine. Signed in 2015 by Iran and several world po [\dots]''

\paragraph{Turn 2.}
\emph{Good search results showing Trump's history with Iran - he withdrew from the nuclear deal in 2018 and relations remain severely strained. Let me look deeper at the Wikipedia article on Trump's withdrawal from the JCPOA to understand the details better, and then search for any more current information about Trump's foreign policy positions or any diplomatic overtures being made.}

\noindent$\triangleright$ \texttt{lookup\_url}(\path{https://en.wikipedia.org/wiki/United_States_withdrawal_from_th}); belief $p{=}0.12$ (low confidence); ``Wikipedia confirms Trump's 2018 JCPOA withdrawal was driven by opposition to Iran's regional influence and ballistic missile program. The agreement's collapse led to Iran resuming nuclear activities, and US-Iran relations deteriorated to near zero diplomacy.''; for: Trump withdrew from JCPOA in October 2017-2018; Iran resumed nuclear program after US withdrawal; US imposed devastating sanctions; No diplomatic engagement on nuclear talks after 2018; against: None identified yet; uncertain: Current campaign positions; Whether Trump would change approach if elected 2024; Geopolitical events that might force negotiations

\noindent$\triangleleft$ \texttt{\{"error": "page rejected by leak filter (llm)"\}}

\paragraph{Turn 3.}
\emph{The URL lookup was rejected by the leak filter. Let me search for more current information about Trump's foreign policy positions on Iran, particularly any recent statements or campaign positions that might indicate whether he's open to diplomacy with Iran.}

\noindent$\triangleright$ \texttt{web\_search}(``Trump 2024 campaign Iran policy foreign position''); belief $p{=}0.1$ (low confidence); ``Historical evidence shows Trump took hardline stance against Iran, withdrawing from nuclear deal and imposing maximum pressure sanctions. No evidence of openness to diplomatic engagement in search results so far.''; for: Withdrew from JCPOA 2018; Implemented maximum pressure campaign; Iran's nuclear program advanced after US withdrawal; Trump administration killed Iraq peace talks; against: None identified yet; uncertain: Current campaign rhetoric; Whether Trump would moderate position if elected; Geopolitical catalysts that might force negotiations

\noindent$\triangleleft$ 1 result(s): \emph{The Impact of the U.S. Presidential Election on US-Iran Relations: American Iranian Council} (us-iran.org): ``The Impact of the U.S. Presidential Election on US-Iran Relations August 12, 2024Stephanie Lester By Elliott Morton, AIC Research Associate As the 2024 U.S. presidential election looms just a few short months away, many [\dots]''

\paragraph{Turn 4.}
\emph{I have one web search remaining according to the count. The search result shows that Iran is looking toward a potential Harris victory as an opportunity for rapprochement, which suggests they don't expect Trump to be open to negotiations. However, I should try to get more detailed information about Trump's actual campaign positions or any statements he's made about Iran diplomacy. Let me try the lookup\_url on that American Iranian Council page to get more details about Trump's stated Iran policy, if the leak filter allows it this time.}

\noindent$\triangleright$ \texttt{lookup\_url}(\path{http://www.us-iran.org/news/uselections}); belief $p{=}0.1$ (low confidence); ``Iranian officials explicitly mentioned considering 2024 candidates but specifically noted Harris was not mentioned in their debates, suggesting they view Trump as unlikely to engage diplomatically with Iran. This aligns with his historical maximum pressure campaign approach.''; for: Iran presidential debates in 2024 didn't mention Biden/Harris; Iran considers Harris victory as window for rapprochement; Trump's maximum pressure campaign 2018-2020; against: None identified yet; uncertain: Whether Trump would have a different approach in second term; Any unexpected geopolitical catalysts; Whether Trump would prioritize other foreign policy issues over Iran

\noindent$\triangleleft$ \texttt{\{"error": "page rejected by leak filter (llm)"\}}

\paragraph{Turn 5.}
\emph{The lookup\_url is being blocked by the leak filter. I have one more web search available. Let me try to find Trump's specific statements or campaign positions on Iran. Even if I don't get definitive information, I should have enough evidence to make a reasonable estimate. The evidence I have: 1. Trump withdrew from JCPOA in 2018 2. Trump implemented a "maximum pressure" campaign on Iran 3. US-Iran relations are at historic lows 4. Iranian officials appear to view Trump as unlikely to engage diplomatically 5. No evidence in searches shows Trump has signaled openness to Iran diplomacy Given that Trump's first term was defined by maximum pressure and withdrawal from diplomatic agreements, and I'm not finding any evidence of a policy shift in my searches, I think the probability is quite low. However, foreign policy can be unpredictable - geopolitical events could force negotiations. Let me try one more search for any recent Trump statements on Iran, then I'll need to submit my estimate.}

\noindent$\triangleright$ \texttt{web\_search}(``Trump Iran sanctions nuclear negotiations 2024''); belief $p{=}0.08$ (medium confidence); ``All evidence points to extremely low probability. Trump's established foreign policy doctrine is unilateral maximum pressure, not diplomatic engagement with Iran. No credible sources suggest any shift in approach, and Iranian officials don't expect Trump to offer diplomatic openings.''; for: Trump withdrew from JCPOA 2018 citing deal deficiencies; Implemented maximum pressure campaign that crippled Iran economy; Iran resuming nuclear activities after US withdrawal; Iranian 2024 presidential candidates didn't discuss US rapprochement options with Trump; US-Iran relations at historic lows with no diplomatic infrastructure; against: None - foreign policy is inherently unpredictable but no catalysts identified; uncertain: Trump's actual election outcome; Geopolitical events that might force negotiations (e.g. nuclear crisis, regional war)

\noindent$\triangleleft$ 1 result(s): \emph{What was the Iran nuclear deal Trump dumped in search of `better' terms?} (aljazeera.com): ``By Usaid Siddiqui Published On 21 Apr 202621 Apr 2026 United States President Donald Trump has said a nuclear agreement currently being negotiated with Iran will be ``far better'' than the 2015 Joint Comprehensive Plan of [\dots]''

\paragraph{Turn 6.}
\emph{Wait - this is very interesting! The Al Jazeera article from April 21, 2026 (which is before my cutoff of June 11, 2026) mentions that "United States President Donald Trump has said a nuclear agreement currently being negotiated with Iran will be 'far better' than the 2015 JCPOA." This suggests that Trump IS currently (as of April 2026, before my cutoff) negotiating a new nuclear deal with Iran. The article also mentions "Before the initial US-Israeli strikes on Iran at the end of February" and discusses "new demands" from the US and Israel. This changes my assessment significantly! If Trump is actively negotiating a nuclear agreement with Iran as of April 2026, then a diplomatic meeting by June 30, 2026 becomes much more likely. However, the article cuts off mid-sentence about "On Sunday, Iranian President Maso" so I can't see the full details. But this is a significant finding that suggests diplomatic engagement IS happening. Let me try to lookup this URL to get the full details before my cutoff, since this could be the key information I need.}

\noindent$\triangleright$ \texttt{lookup\_url}(\path{https://www.aljazeera.com/news/2026/4/21/what-was-the-iran-nuc}); belief $p{=}0.75$ (medium confidence); ``CRITICAL NEW EVIDENCE: Al Jazeera article dated April 21, 2026 (before my cutoff) reports Trump said a nuclear agreement 'currently being negotiated with Iran will be far better' than the 2015 JCPOA. This indicates active nuclear negotiations are underway as of April 2026. The article also references US-Israeli strikes in February 2026, suggesting high-stakes context that could drive Trump toward a diplomatic solution.''; for: Al Jazeera April 21, 2026: Trump says nuclear agreement 'currently being negotiated with Iran' will be 'far better' than 2015 JCPOA; US and Israel making new demands including zero uranium enrichment; Trump administration actively engaged in nuclear deal negotiations; Context of Feb 2026 US-Israeli strikes on Iran may have created urgency for diplomatic resolution; against: Previous negotiations collapsed when Trump withdrew in 2018; Iran escalated to 6,000 new centrifuges in Nov 2024; Negotiations could still fail before June 30, 2026; uncertain: Are the negotiations at a stage where a physical meeting is being planned?; Will negotiations fail before reaching agreement?; Will either party block a face-to-face meeting?

\noindent$\triangleleft$ ``- **Trump's Stance on Iran Nuclear Deal**: Donald Trump has stated that any new nuclear agreement with Iran will be "far better" than the 2015 Joint Comprehensive Plan of Action (JCPOA), which he withdrew from in 2018. He seeks stronger terms, including additional restrictions on Iran's nuclear and ballistic missile programs, and an end to its support for regional armed groups (search\_1\_J). - **Current Negotiations and Ceasefire**: As of April 2026, there is uncertainty regarding the progression of a second round of talks in Islamabad, Pakistan, with a ceasefire between the US-Israel and Iran [\dots]''

\paragraph{Turn 7.}
\emph{Excellent! The lookup gave me the full article summary. Key findings: 1. Trump said a nuclear agreement "currently being negotiated with Iran will be far better than the 2015 JCPOA" 2. There's uncertainty about "a second round of talks in Islamabad, Pakistan" 3. NO CONFIRMED PHYSICAL MEETING involving Trump directly - all negotiations are indirect or third-party mediated 4. No reported evidence or announcement of Trump personally meeting with Iranian representatives by the June 11, 2026 cutoff This is important - there ARE negotiations happening, but they're being conducted through third parties (like Islamabad), not through direct face-to-face meetings between Trump and Iranian representatives. The criteria specifically require: - Trump must be PHYSICALLY PRESENT at the meeting - He must ACT as a negotiator representing the US - The meeting must be in-person and publicly acknowledged - Remote meetings or meetings with mediators don't count So while negotiations are happening, the evidence suggests they're being mediated through Pakistan or other intermediaries, not a direct Trump-Iran meeting. However, there's still uncertainty: - The negotiations are happening "as of April 2026" with uncertainty about a second round in Islamabad - The ceasefire is "nearing its end" as of April - The article doesn't say negotiations will FAIL - just that they're uncertain - A direct meeting could still be planned in May or early June Given that active negotiations are happening but no confirmed direct meeting exists, and considering Trump's historical preference for unilateral actions rather than multilateral diplomacy, I should lower my estimate from 0.75. The fact that all negotiations appear to be through intermediaries rather than direct meetings is a strong indicator. But wait - I should be conservative. Even if the probability is low, negotiations are ACTIVE and there's no deadline in the criteria saying negotiations must conclude, only that a meeting must occur. A breakthrough meeting could theoretically happen. Let me settle on a probability between 0.15-0.25, reflecting that while negotiations are happening, the lack of a confirmed direct Trump meeting and the mediated nature of current talks make a direct meeting by June 30 unlikely but not impossible.}

\noindent$\triangleright$ \texttt{submit}($p=0.2$); ``Evidence shows active nuclear negotiations between the US and Iran as of April 2026, with Trump stating a deal would be 'far better' than the 2015 JCPOA. However, all available information indicates these negotiations are conducted through third-party mediators (Pakistan/Islamabad), not through direct face-to-face meetings. The criteria require Trump to be PHYSICALLY PRESENT and actively participating as a US representative, which current evidence does not support. While negotiations could theoretically culminate in a direct meeting in May/early June, Trump's historical preference for maximum pressure over diplomacy, the mediated nature of current talks, and the absence of any announcement of a direct meeting by the June 11 cutoff make this unlikely. The 0.20 probability reflects active negotiations with non-trivial (though low) probability of a breakthrough direct diplomatic meeting.''; belief $p{=}0.2$ (medium confidence); ``Active nuclear negotiations confirmed as of April 2026, but mediated through third parties (Pakistan), not direct Trump-Iran meetings. Evidence points strongly toward 'No' but remaining diplomatic context prevents near-zero probability.''; for: Al Jazeera April 21, 2026: Trump says nuclear agreement 'currently being negotiated with Iran'; US seeking zero enrichment and missile restrictions from Iran; Context of Feb 2026 US-Israeli strikes on Iran may create urgency; against: No confirmed physical Trump-Iran diplomatic meeting reported by June 11 cutoff; All negotiations appear to be mediated through third parties (Islamabad, Pakistan); Iran refuses to eliminate 440kg enriched uranium stockpile at 60\% purity; Trump's historical precedent of preferring unilateral maximum pressure over diplomacy; uncertain: Whether negotiations succeed and lead to direct Trump meeting; Geopolitical catalysts that might force negotiations; Trump's preference for direct deals vs. mediated negotiations

\noindent$\triangleleft$ \texttt{\{"status": "submitted", "probability": 0.2\}}

\subsection*{Claude Opus 4.5, same question: ``Will Donald Trump have a diplomatic meeting with Iran by June 30?''}
Evidence-based setting. Cutoff 2026-06-11; resolved NO (crowd $0.11$).
Submitted $0.35$; Brier $0.12$.

\paragraph{Turn 1.}
\noindent$\triangleright$ \texttt{web\_search}(``Trump Iran diplomatic meeting negotiations 2025 2026'')

\noindent$\triangleleft$ 0 result(s) (\texttt{web\_searches\_remaining}: 2)

\paragraph{Turn 2.}
\emph{Let me search with more specific terms about recent US-Iran diplomatic developments.}

\noindent$\triangleright$ \texttt{web\_search}(``US Iran talks 2025 Trump administration meeting'')

\noindent$\triangleleft$ 1 result(s): \emph{2025 Iran-United States negotiations} (en.wikipedia.org): ``2026 negotiations On October 13, 2025, Trump stated that the United States was ready to reach an agreement with Iran. He expressed optimism that an agreement could be achieved. In response, Iranian foreign minister Araqchi said Tehran would consider any proposal from the United States that was ``fair and balanced,'' though he noted that no concrete framework for talks had yet been presented and that messages between the two sides were exchanged through intermediaries.[197] Despite the conciliatory language from Trump [\dots]''

\paragraph{Turn 3.}
\emph{This is very informative. Let me look up the Wikipedia page for more detailed information on the status of these negotiations.}

\noindent$\triangleright$ \texttt{lookup\_url}(\path{https://en.wikipedia.org/wiki/2025_Iran%E2%80%93United_States_})

\noindent$\triangleleft$ \texttt{\{"error": "page rejected: contains dates after cutoff (2026-06-18)"\}}

\paragraph{Turn 4.}
\emph{The page was rejected because it contains post-cutoff information. Let me search for more specific recent news about Trump-Iran meeting developments.}

\noindent$\triangleright$ \texttt{web\_search}(``Trump Iran direct meeting summit June 2026'')

\noindent$\triangleleft$ 0 result(s) (\texttt{web\_searches\_remaining}: 0)

\paragraph{Turn 5.}
\emph{I've used my search budget. Let me analyze what I found: From search\_1\_result\_0, I learned: 1. As of October 13, 2025, Trump stated the US was ready to reach an agreement with Iran 2. On February 6, 2026, the US and Iran held **indirect** talks in Muscat, Oman, mediated by Oman's foreign minister 3. Crucially: "Iran's foreign minister Abbas Araqchi did not meet face to face with any members of the U.S. envoy and the [\dots]}

\noindent$\triangleright$ \texttt{submit}($p=0.35$); ``Based on the evidence from search results: (1) As of February 6, 2026, US-Iran talks were still being conducted indirectly through Omani mediation, with Iranian FM Araqchi not meeting face-to-face with US envoys; (2) Trump expressed willingness to reach agreement in October 2025, but "significant barriers" still exist; (3) The resolution requires Trump to be physically present at the meeting. While both sides have ag [\dots]''

\noindent$\triangleleft$ \texttt{\{"status": "submitted", "probability": 0.35\}}

}

\section{Reproducibility}
\label{app:repro}
The full training stack is open source (\texttt{verifiers},
\texttt{prime-rl}) and runs on self-managed GPUs; we used the hosted
service as a cost and convenience choice, as prior work used managed
fine-tuning services. The environment is public on the Prime Intellect hub
(\url{https://app.primeintellect.ai/dashboard/environments/yafifi/prime-forecast}); per-rollout
records of every evaluation, all figures' source data, and run
configs are in the repository
(\url{https://github.com/afifi-yusuf/prime-forecast}).

\end{document}